\documentclass[letterpaper, 10 pt, conference]{ieeeconf}  

\IEEEoverridecommandlockouts                              

\usepackage{bm}
\usepackage{url}
\usepackage{graphicx} 
\usepackage{mathptmx} 
\usepackage[mathscr]{euscript}  
\usepackage{amsmath} 
\usepackage{amssymb}  
\usepackage{multirow}
\usepackage[binary-units=true,load=prefixed]{siunitx}
\usepackage{xcolor}
\usepackage{times}
\usepackage{booktabs}
\usepackage[font=scriptsize]{caption}
\usepackage[font=scriptsize]{subcaption}

\usepackage{hyperref}
\hypersetup{
    colorlinks=true,
    linkcolor=blue,
    filecolor=magenta,      
    urlcolor=cyan,
}

\renewcommand{\baselinestretch}{0.97}

\newcommand{\secref}[1]{Sec.~\ref{#1}}

\renewcommand{\eqref}[1]{Eq.~(\ref{#1})}
\newcommand{\figref}[1]{Fig.~\ref{#1}}
\newcommand{\tabref}[1]{Tab.~\ref{#1}}

\newcommand\tab[1][0.5cm]{\hspace*{#1}}

\DeclareSIUnit\degm{deg/m}
\newcommand{\rot}[1]{\rotatebox{90}{#1}}

\title{\LARGE \bf
Vision-based Autonomous Landing in Catastrophe-Struck Environments
}

\author{Mayank Mittal$^{*}$ \and Abhinav Valada$^{*}$ \and Wolfram Burgard 
\thanks{$^*$These authors contributed equally. All authors are with the Department of Computer Science, University of Freiburg, Germany. This work has partly been supported by the Federal Ministry of Education and Research of Germany through the project FOUNT2.}
}

\begin{document}

\maketitle
\thispagestyle{empty}
\pagestyle{empty}

\begin{abstract}
Unmanned Aerial Vehicles (UAVs) equipped with bioradars are a life-saving technology that can enable identification of survivors under collapsed buildings in the aftermath of natural disasters such as earthquakes or gas explosions. However, these UAVs have to be able to autonomously land on debris piles in order to accurately locate the survivors. This problem is extremely challenging as the structure of these debris piles is often unknown and no prior knowledge can be leveraged. In this work, we propose a computationally efficient system that is able to reliably identify safe landing sites and autonomously perform the landing maneuver. Specifically, our algorithm computes costmaps based on several hazard factors including terrain flatness, steepness, depth accuracy and energy consumption information. We first estimate dense candidate landing sites from the resulting costmap and then employ clustering to group neighboring sites into a safe landing region. Finally, a minimum-jerk trajectory is computed for landing considering the surrounding obstacles and the UAV dynamics. We demonstrate the efficacy of our system using experiments from a city scale hyperrealistic simulation environment and in real-world scenarios with collapsed buildings.
\end{abstract}


\section{Introduction}
\label{sec:Introduction}

Search and rescue operations for finding victims in collapsed buildings is an extremely time critical and dangerous task. There are several triggers for buildings to collapse including gas explosions, fires as well as natural disasters such as storms and earthquakes. The current paradigm followed during the response and recovery phase of the disaster management cycle is to first conduct a manual inspection of the damaged structures by disaster response teams and firefighters, followed by actions to search for victims using bioradars and thermal cameras. However, there are often several inaccessible areas that take anywhere from a few hours to days to reach, which not only endangers the lives of the trapped victims but also the rescue team due to the inherent instability of the rubble piles.

These factors have increased the interest in employing Unmanned Aerial Vehicles (UAVs) for reconnaissance operations due to their agile maneuverability, fast deployment and their ability to collect data at high temporal frequencies. Typically, in addition to optical sensors such as thermal cameras, ground penetrating radars such as bioradars that can detect movements in the internal organs of humans such as the lungs and heart are used for bioradiolocation~\cite{soldovieri2012feasibility}. More recently, our partner researchers have miniaturized a bioradar~\cite{shi2017design} capable of being mounted as payload on small-sized UAV. However, due to the large difference in the electromagnetic impedance between the air and collapsed structures, the antenna of the bioradar should be in contact with the surface of the rubble in order to obtain accurate measurements. Therefore, the UAV should be capable of reliably detecting safe landing spots in collapsed structures and autonomously perform the landing maneuver. This problem is extremely challenging even for an expert human operator as almost no prior knowledge of the environment such as digital surface maps can be leveraged. Moreover, as often communication infrastructures are also damaged during natural disasters, all the computation has to be performed on-board on a resource constrained embedded computer.

\begin{figure}
\footnotesize
\centering
\setlength{\tabcolsep}{0.1em} 
\begin{tabular}{p{0.5cm} p{0.42\textwidth}} 
\rot{(a) Hyperrealistic Simulation} & \includegraphics[width=\linewidth]{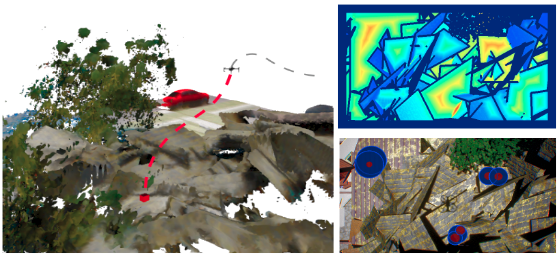} \\
\rot{(b) Real World Scenario} & \includegraphics[width=\linewidth]{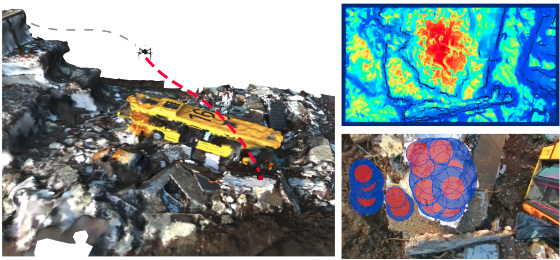} \\
\end{tabular} 
\caption{Illustration of the textured 3D volumetric reconstruction of collapsed buildings showing the minimum-jerk trajectory to the detected safe landing site. The corresponding full costmap computed by our landing site detection algorithm is shown on the right along with the dense candidate landing sites overlaid on the visual image of the scene. The blue circles show the detected safe landing spots and the enclosed red circles indicate the area occupied by the UAV. The circles appear as different sizes due to the 2D projection on planes at different distances from the camera.}
\label{fig:landing_cover}
\vspace{-0.5cm}
\end{figure}

Most existing work on landing site detection employ specific markers or patterns that can be identified by a UAV as the landing site, such as the alphabet 'H' on helipads or fiducial markers~\cite{lee_vtol}. These approaches require the landing site to be predetermined and are not employable in unstructured or unknown environments. Other existing approaches that only employ planarity constraints~\cite{bosch2006autonomous} are insufficient for our application as our UAV is required to land on collapsed buildings that appear as rubble piles. Due to the nature of these collapsed structures, the algorithm should be able to handle a multitude of visual terrain features and natural or man-made obstacles that the UAV might encounter in a post disaster struck environment. Moreover, a critical requirement being that the landing site detection algorithm has to run online on a low-power embedded computer along with other autonomy packages for state estimation and mapping.

In this paper, we present a solution to the aforementioned problem through a robust vision-based autonomous landing system that can detect safe landing sites on rubble piles from collapsed buildings in real-time and perform the landing maneuver autonomously to the landing region. Our proposed algorithm assesses the risk of a landing spot by evaluating the flatness, inclination and the orientation of the surface normals using the depth map from a stereo camera, in addition to considering the confidence of the depth information inferred and the energy required to land on the detected spot. A set of dense candidate landing sites are first estimated locally by computing a weighted sum of the costmaps for each of the hazard factors, followed by optimizing over the global area explored by the UAV using a clustering based approach. We then project the landing sites on to a textured 3D volumetric reconstruction of the area which is computed real-time on-board of the UAV. Once the final landing site has been chosen, the system computes a minimum-jerk trajectory considering the nearby obstacles as well as the UAV dynamics and executes the landing maneuver.

We evaluate our proposed system using extensive experiments in a hyperrealistic city scale simulated environment and in real-world environments with collapsed buildings as is exemplary shown in \figref{fig:landing_cover}. To the best of our knowledge, this is the first such system capable of landing on rubble piles in catastrophe-struck environments. Although we demonstrate the utility of our system for autonomous landing on collapsed buildings, it can also be employed for landing in planned or emergency situations such as when the UAV has low battery or has lost communication to the ground station.

\section{Related Work}
\label{sec:relatedWork}

In the last decade, a wide range of vision-based landing sites detection approaches have been proposed for UAVs. These techniques can be categorized into methods that either employ fiducial markers for landing at known locations or assess various surface and terrain properties for landing in an unknown environment. In the first category of approaches, markers are detected on the basis of their color or geometry using classical image features and then the relative pose of the UAV is estimated from these extracted feature points. Over the years, several types of fiducial markers have been proposed for this purpose including point markers~\cite{mebarki2015autonomous}, circle markers~\cite{lange2009vision}, H-shaped markers~\cite{hu2015fast} and square markers~\cite{chaves2015neec}. More recently, advancement in techniques for robust tracking of fiducial markers using IR LEDs~\cite{wenzel2011automatic} has also been made. While the usage of fiducial markers for landing purposes is reliable and efficient, they are usable only when the desired landing spots are known in advance such as for landing on ship decks~\cite{grocholskyrobust} or a moving mobile robot platform~\cite{falanga2017vision}.

More relevant to our work are the approaches which estimate safe landing sites in unknown or unstructured environments. Forster~\textit{et al.}~\cite{forster2015continuous} propose an approach that builds a 2D probabilistic robot-centric elevation map from which landing spots are detected over regions where the surface of the terrain is flat. Templeton~\textit{et al.}~\cite{templeton2007autonomous} present a terrain mapping and landing system for autonomous helicopters that computes the landing quality score based on the linear combination of the angle of the best-fit horizontal plane, the plane fit error, and other appearance factors. Concurrently, the authors in~\cite{garcia2002towards} propose an approach for safe landing of an autonomous helicopter in regions without obstacles by utilizing a contrast descriptor and correlation function to detect obstacles under the assumption that the boundaries of obstacles have high contrast regions. A stereo vision-based landing site search algorithm is presented in~\cite{park2015landing}, in which a performance index for landing is computed considering the depth, flatness, and energy required to reach a specific site. Desaraju~\textit{et al.}~\cite{desaraju2014vision} employ an active perception strategy utilizing Gaussian processes to estimate feasible rooftop landing sites along with the landing site uncertainty as assessed by the vision system.

Bosch~\textit{et al.}~\cite{bosch2006autonomous} introduce an approach that considers a sequence of monocular images for robust homography estimation in order to identify horizontal planar regions for landing as well as for estimating the motion of the camera. In~\cite{cheng2010real}, the authors propose a similar technique for camera motion estimation and detection of multiple planar surfaces in complex real-world scenes. While in~\cite{theodore2006flight}, Theodore~\textit{et al.} presents an approach in which they first create a stereo range map of the terrain and then choose the landing point based on the surface slope, roughness and the proximity to obstacles. Johnson~\textit{et al.}~\cite{johnson2002lidar} propose a Lidar-based approach in which an elevation map is computed from Lidar measurements, followed by thresholding the regions based on local slope and roughness of the terrain. Most recently, Hinzmann~\textit{et al.}~\cite{hinzmann2018free} present a landing site detection algorithm for autonomous planes in which they first employ a binary random forest classifier to select regions with grass, followed by 3D reconstruction of the most promising regions from which hazardous factors such as terrain roughness, slope and the proximity to obstacles that obscure the landing approach are computed to determine the landing point.

In contrast to the aforementioned techniques, the approach presented in this paper detects safe landing sites on collapsed buildings which often appear as rubble piles therefore it employs fine-grained terrain assessment considering a wide range of hazardous factors at the pixel-level. By first estimating dense candidate landing sites in a local region, followed by a global refinement, the approach is able to run online along with other state estimation and mapping processes on a single on-board embedded computer mounted on a UAV. 

\section{Technical Approach}
\label{sec:technicalApproach}

\begin{figure*}
\centering
\includegraphics[width=	\linewidth]{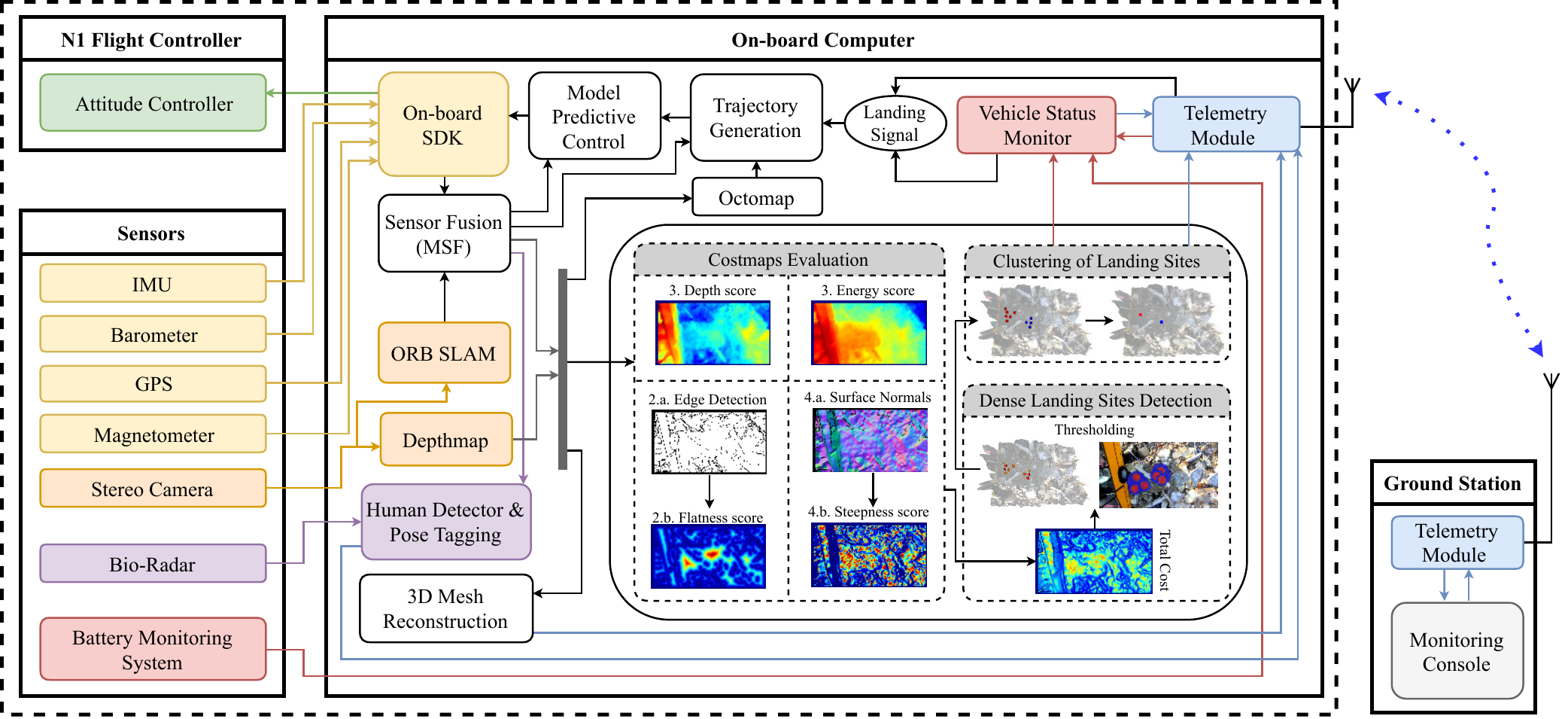}
\caption{Overview of our autonomous landing system. We use the DJI M100 quadrotor with the N1 flight controller and NVIDIA Jetson TX2 for the on-board computer. We equip our quadrotor with a ZED stereo camera for acquiring depth information. All our mapping, localization and landing site detection algorithms run online on the TX2.}
\label{fig:overview_system}
\vspace{-0.5cm}
\end{figure*}

In this section, we first briefly describe the overall architecture of our autonomous landing system, followed by detailed descriptions of each of the constituting components. \figref{fig:overview_system} shows an overview of our proposed system. We estimate the current pose of the UAV by fusing raw data from the inertial sensors with GPS measurements and the poses obtained from stereo ORB-SLAM2~\cite{raulOrbSLAM} using an extended Kalman filter (EKF) as described in~\secref{sec:stateEstimation}. Accurate estimation of the UAV's pose is an essential requirement for various modules including volumetric mapping, localization of the detected humans from the bioradar measurements and for detecting landing sites using our proposed algorithm. The autonomous landing protocol detailed in~\secref{sec:landingDetection} can be subdivided into three stages. In the first stage, we evaluate the costmaps for various metrics such as terrain flatness, steepness, depth accuracy and energy consumption information in the local camera frame. We then infer a set of dense candidate landing sites from the combined costmaps, followed by employing a nearest neighbor filtering and clustering algorithm to obtain a sparser set of unique landing sites in the global frame.

In our system, we generate two different 3D representations of the environment - an OctoMap~\cite{hornung13auro} which serves as a light voxel-based 'internal' map for the UAV for trajectory planning and a 3D mesh reconstruction~\cite{oleynikova2017voxblox} which is transmitted to the ground station for analysis and verification by a human operator. We describe the mapping procedure in~\secref{sec:mapping}. The UAV also transmits the vehicle status information, landing sites detected in the explored area and poses of any detected humans. For the sake of generality, we express the situations for planned (normal operations) and forced landings (emergency situations) using a common landing signal. In case of a planned operation, the operator selects a specific site to land from the list of candidate sites detected in the region. While, in case of emergencies, the UAV chooses a landing site based on whether it needs to land quickly (low battery) or at a location closest to the remote station (loss of communication signal). Once the landing signal has been transmitted/generated, a minimum-snap trajectory is planned on-board to safely land at the selected site as described in \secref{sec:trajectoryEstimation}. Finally, the waypoints indicating the planned trajectory are sent to a position-based model predictive controller which sends the actuation commands to the flight controller of the UAV.

\subsection{State Estimation}
\label{sec:stateEstimation}

For autonomous operation of UAVs, it is crucial to reliably estimate the vehicle's position in the world. Although GPS provides a straightforward solution, it is highly unreliable in complex cluttered or confined areas as well as in other GPS-denied environments. Therefore, we use ORB-SLAM2~\cite{raulOrbSLAM} to estimate the pose of the vehicle from the image stream of a downward-facing stereo camera mounted on the UAV. The algorithm extracts ORB (oriented FAST and rotated BRIEF) features from the input frames and performs motion-only bundle adjustment for tracking the extracted features. It utilizes a bag-of-words representation during tracking to match features and runs efficiently even on embedded computers. Since we use a stereo configuration in our system, the pose estimated from ORB-SLAM2 is in absolute scale, therefore no sensor fusion for scale correction is necessary. However, to further improve the accuracy of the estimated pose, we fuse the output from the ORB-SLAM2 system with data from the on-board inertial measurement unit (IMU), barometer and GPS using the multi-sensor fusion (MSF) module~\cite{Lynen_MSF} which utilizes an extended Kalman Filter. The aforementioned sensors are pre-calibrated using the Kalibr toolbox~\cite{furgale2013unified}. This state estimation system provides precise localization information even in complex environments with collapsed buildings.

\subsection{Landing Site Detection}
\label{sec:landingDetection}

\begin{figure*}
\centering
\includegraphics[width=0.95\linewidth]{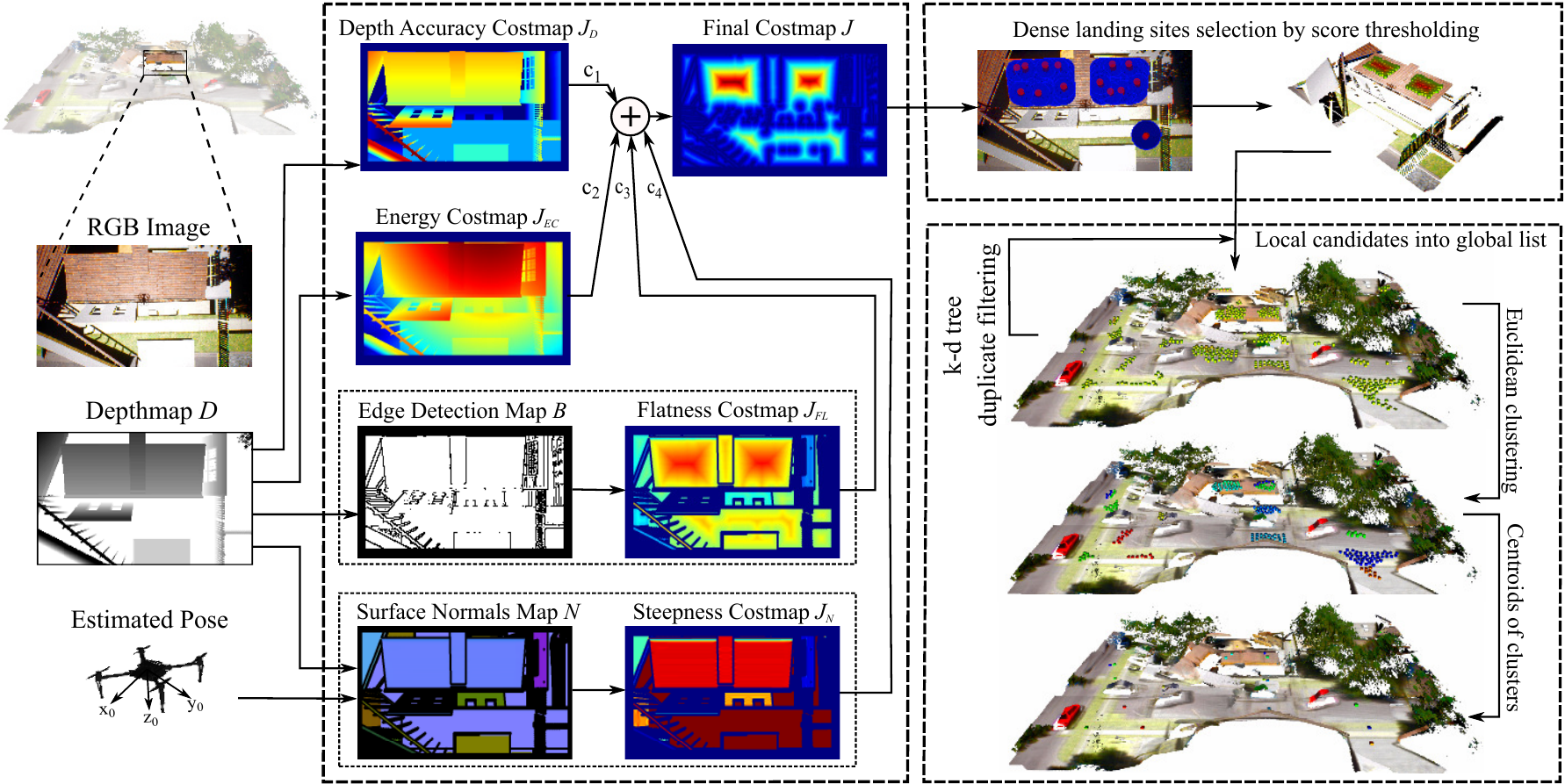}
\caption{Overview of our landing site detection algorithm. The figure illustrates the various costmaps for hazard estimation. \textit{Scale:} Red indicates high score while blue indicates lower score. The detected landing sites are projected on to a volumetric 3D reconstructed mesh of the environment.}
\label{fig:landing_algo}
\vspace{-0.5cm}
\end{figure*}

The criteria for the selection of candidate landing sites is to locate regions from aerial imagery that are reasonably flat, within the range of the accepted slope, free of obstacles and large enough for the UAV to land on. Quantifying each of these requirements through a costmap makes our approach generic to the various structures that the UAV may encounter in the aftermath of catastrophic events, as well as in other emergency situations. Using the estimated current pose of the UAV and the depth map obtained from synchronized stereo images, the key costmaps that we compute in our algorithm are described in the following sections. \figref{fig:landing_algo} provides an overview of our landing site detection algorithm. We denote the minimum and the maximum range for which the sensor data is valid by $d_{min}$ and $d_{max}$, and the depth map obtained from the stereo camera by $D$.

\subsubsection{Confidence in Depth Information $J_{DE}$}

In~\cite{nguyen2012kinect}, Nguyen~\textit{et al.} empirically derived a noise model for an RGB-D sensor according to which the variance in axial noise distribution of a single ray measurement varied as a squared function of the depth measurement in the camera frame. More specifically, when the UAV is operating at a high altitude, inaccurate depth information may be obtained for objects closer to the ground. In order to encode this confidence on the depth information, we evaluate the costmap function $J_{DE}$ such that $J_{DE}(p) \in [0, 1]$ is the score given to each pixel $p=(x, y)$ in the depth map such that
\begin{equation}
J_{DE}(p) = 1 - \frac{D(p)^2 - \min(D^2)}{\max(D^2)}.
\end{equation}

\subsubsection{Flatness Information $J_{FL}$}

From the depth map of the topography containing ground obstacles, the flatness of an area can be represented by the portion in the map having the same depth. The size of this area can then be determined by inscribing a circle in every portion of the image with the same depth level and selecting the location which has the largest diameter. We evaluate the flatness information using the aforementioned analogy. By applying a Canny edge detector over the depthmap, we obtain a binary image where non-zero elements represent the edges for depth discontinuities. In order to find the inscribed circle, we apply a distance transform which assigns a number to each pixel depending on the distance of that pixel to the nearest nonzero pixel of the binary edge map. For each pixel $p=(x,y)$ in the image plane, the Euclidean distance transform of a binary image $B$, is defined as
\begin{equation}
di(B, p) = min\Big\{\sqrt{(p-q)^T (p-q)} \Big{|} B(q) = 1 \Big\}.
\end{equation}
Using this operator, we calculate the flatness score as
\begin{equation}
J_{FL}(p) = di(Canny(D), p).
\end{equation}

It can be inferred that the flatter areas in the depth map are given higher scores in the evaluated flatness map $J_{FL}$.

\subsubsection{Steepness Information $J_N$}

Another criteria to measure the quality of a landing site is based on the steepness of the area around the region. We quantify the steepness by computing the surface normals from the generated depth map. However, while estimating this information online, it is essential to account for the deviation in the calculated normals due to the orientation of the UAV as the depth map is represented in the camera frame. In order to account for this factor, we generate a point cloud from the depth map and transform it into the global frame. Using the transformed depth information, we estimate the surface normals using the average 3D gradients algorithm to obtain a normals map $N$. Compared to the covariance based method, the average 3D gradients approach computes the normals significantly faster and the results are comparable to a 30-nearest neighborhood search in the covariance method~\cite{nakagawa3dcv}. For each pixel in the calculated surface normals map $N$, we evaluate  deviation of the normalized surface normal $\hat{n}$ with respect to the z-axis in the global frame using the vector dot product as
\begin{equation}
\theta = cos^{-1}(\hat{n}.\hat{z}).
\end{equation}
The steepness score for each pixel $p$ is then given by
\begin{equation}
n(p) = \exp\Bigg\{-\frac{\theta^2}{2\theta_{th}^2}\Bigg\},
\end{equation}
where we set $\theta_{th}$ to $15^o$ in this work, which is the maximum tolerable slope that our UAV can perch on safely. 

\subsubsection{Energy Consumption Information $J_{EC}$}

Often, there are several flat areas where the UAV could potentially land and in some cases, it might be desirable to land on a site that consumes lesser energy to navigate to. In order to account for this factor, we compute the energy consumption required to follow a safe trajectory to a landing site at pixel $p$ as
\begin{equation}
J_{EC}(p) = \int_{t_{o}}^{t_{f}} P(t) dt,
\end{equation}
where $t_0$ and $t_f$ are the time of flight for the path to reach location $p$ and $P(t)$ is the instantaneous battery power. However, computing a costmap by evaluating this integral is a computationally expensive task since a trajectory for the UAV would need to be computed for each pixel. However, the battery power consumed by the UAV is directly related to the amount of energy required to reach the location of that pixel~\cite{park2015landing}. Therefore, we can approximate the aforementioned integral operation by computing the Euclidean distance in the 3D space between the UAV and the location of that point relative to body frame of the UAV. Following this approximation, we assign the value computed for each pixel to obtain the costmap $J_{EC}$.

\subsubsection{Dense Landing Sites Detection}

After evaluating the individual costmaps, we perform min-max normalization over the depth accuracy, flatness and the energy consumption costmaps to scale their values to the same range and remove any biases due to unscaled measurements in the evaluated costs. We then take a weighted sum of the scores assigned to each pixel in their respective costmaps and calculate a final decision map $J$ given by
\begin{equation}
J = c_1 J_{DE} + c_2 J_{FL} + c_3 J_{N} + c_4 J_{EC},
\end{equation}
where $c_1$, $c_2$, $c_3$ and $c_4$ are weighting parameters for each map with the constraints $c_i \in [0,1]$, $\forall i \in \{1, 2, 3, 4\}$ and $c_1 + c_2 + c_3+ c_4 =1$. 

The sites with scores above a certain threshold are considered as candidate landing sites in the local input frame. We perform further filtering of the landing site by evaluating whether the area available around each site is large enough for a UAV to land on. We achieve this by comparing the flatness score of the site to the pixel-wise size of the UAV obtained by projecting the UAV on to the image plane using a pinhole camera model and the depth information of the landing site. Once the filtering has been performed, the locally detected candidate sites are then forwarded to the next stage for aggregation into a global list of detected sites.

\subsubsection{Clustering of Landing Sites}

As we perform landing site detection on frame-to-frame basis, the same landing site detected on the image frame maybe detected in different frames depending on the motion of the UAV. Therefore, in order to account for this factor, we use the depth information and the pose of the UAV to infer the 3D position corresponding to the pixel coordinates of each detected landing site. We then store this location using a k-D tree to efficiently search for a neighboring landing site within a certain distance threshold to an input candidate site. We add the location of the new landing site to the global list only if it currently has no existing neighbors in the list. 

Moreover, in large flat regions, several landing sites might be detected within close proximity of each other. If the entire exhaustive list of detected sites is provided to the human operator, it reduces the reaction time which is critical in rescue situations. In order to alleviate this problem, we apply an agglomerative hierarchical clustering algorithm over the global list of detected landing sites. Clusters are formed on the basis of the euclidean distance between landing sites and the difference in their locations along the z-axis. This yields a sparse set of landing sites with each site location corresponding to the centroid of a cluster. This also helps in overcoming the effect of drift $(d_x, d_y, d_z)$ in the estimated 3D positions of dense landing sites in events such as loop closures. Since the centroid of a cluster $k$ is given by
\begin{equation}
\resizebox{.9\hsize}{!}{$
(x_{c,k}, y_{c,k}, z_{c,k}) = \Bigg(\frac{\sum_{i=1}^{m} x_{i,k} + d_x}{m},\frac{\sum_{i=1}^{m} y_{i,k} + d_y}{m}, \frac{\sum_{i=1}^{m} z_{i,k} + d_z}{m}\Bigg)$},
\end{equation}
the net drift of the resulting centroid is reduced by a factor of the number of dense sites $m$ present in that cluster. We set the cluster distance threshold as a factor of the UAV size.

\subsection{3D Volumetric Mapping}
\label{sec:mapping}

\begin{figure}
\footnotesize
\centering
\setlength{\tabcolsep}{0.1em}
\renewcommand{\arraystretch}{0}
\begin{tabular}{p{0.5cm} p{0.42\textwidth}} 
\rot{(a) OctoMap representation} & \includegraphics[width=\linewidth]{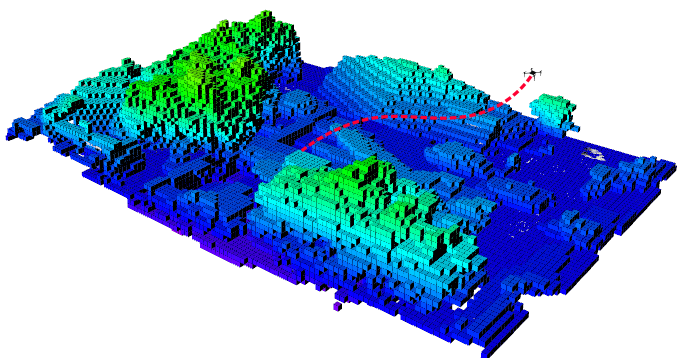} \\ 
\rot{(b) Mesh visualization} & \includegraphics[width=\linewidth]{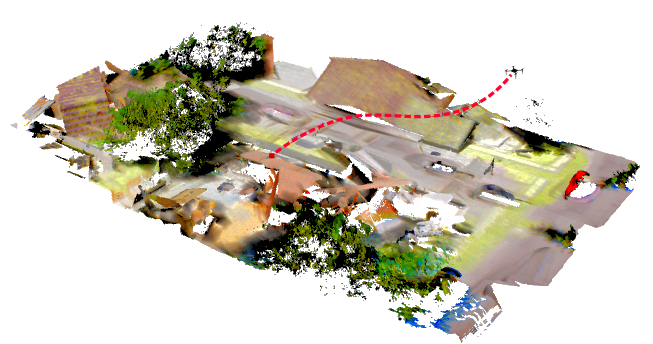} 
 \\
\end{tabular} 
\caption{A minimum-jerk polynomial trajectory is generated to the selected landing site while considering a minimal set of waypoints and a differential flat model of the quadrotor. The on-board planner uses the OctoMap representation for planning a collision free path, while the mesh visualization is created for structural analysis as well as search and rescue planning.}
\label{fig:traj_cover}
\vspace{-0.5cm}
\end{figure}

Exploring the unknown environment consisting of collapsed buildings is one of the foremost tasks for the UAV during the search and rescue operation. The UAV is required to build a map of the environment not only for its navigation but also for the rescue team to remotely assess the situation. In order to create 3D volumetric maps, the UAV primary relies on the depth information from the stereo camera to sense the environment. We generate two different 3D map representations of the environment, an occupancy grid for its navigation and planning, and a textured 3D mesh for visualization. The two maps are generated at different resolutions since a high-resolution map is required for planning a human-lead rescue operation, while a low-resolution map enables faster trajectory planning of collision-free paths for the UAV navigation.\looseness=-1

We use OctoMaps~\cite{hornung13auro} for the internal representation of the environment at a low-resolution (typically $0.5\meter$). OctoMaps use octrees to efficiently create a probabilistic 3D volumetric map of the area and models the environment efficiently. Whereas, for generating a 3D textured mesh, we employ Voxblox~\cite{oleynikova2017voxblox} which is based on Truncated Signed Distance Fields (TSDFs). This framework allows for dynamically growing maps by storing voxels in form a hash table which makes accessing them more efficient compared to an octree. Using Voxblox, we reconstruct a textured mesh from the updated TSDF voxels on the on-board NVIDIA Jetson TX2 and transmit the mesh for analysis by the rescue team.\looseness=-1

\subsection{Landing Trajectory Estimation}
\label{sec:trajectoryEstimation}

When the human operator selects a landing site or when the on-board vehicle status monitor detects the need to land, the UAV plans a trajectory to the selected site and initiates the landing maneuver. To do so, we utilize a minimum-jerk trajectory generator~\cite{richter2016polynomial} with non-linear optimization. The algorithm firsts finds a collision free path to the landing site using RRT*, followed by generating waypoints from the optimal path according to a line-of-sight technique. Using unconstrained nonlinear optimization, it generates minimum-snap polynomial trajectories while considering the minimal set of waypoints and a differential flat model of the quadrotor. This allows the UAV to travel in high speed arcs in obstacle free regions and ensures low velocities in tight places for minimum jerk around corners. \figref{fig:traj_cover} shows the trajectory planned to a safe landing site along with the OctoMap and the textured mesh of the area explored by the UAV.

\section{Experimental Evaluation}
\label{sec:results}

\begin{figure}
  \begin{subfigure}[b]{0.48\columnwidth}
    \includegraphics[width=\linewidth]{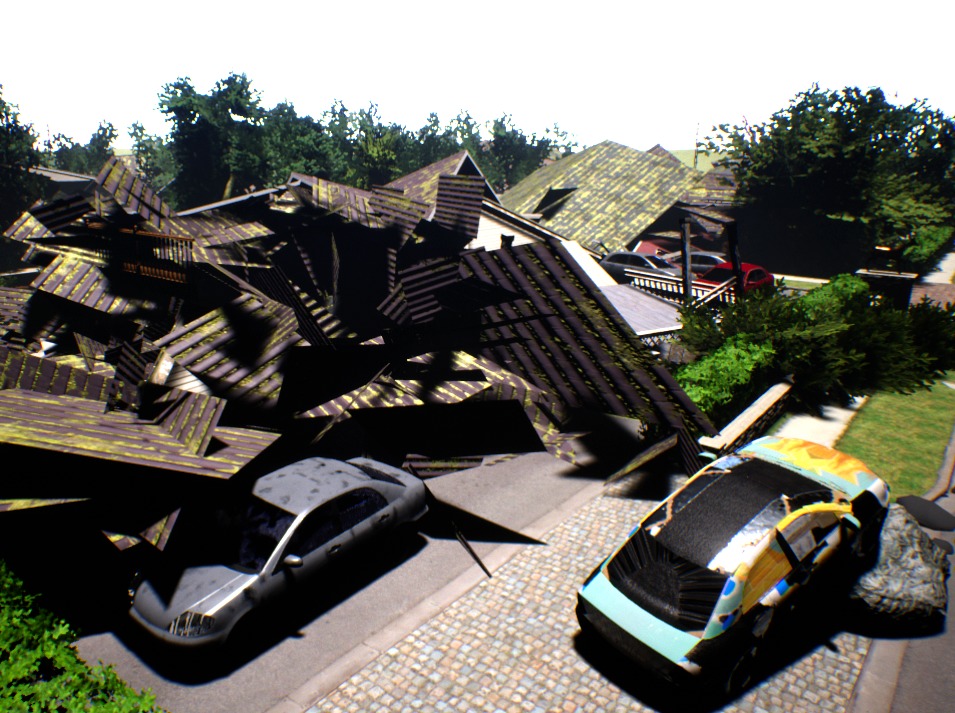}
    \caption{Simulation Environment}
    \label{fig:unreal_envmt}
  \end{subfigure}
  \hfill 
   \begin{subfigure}[b]{0.48\columnwidth}
    \includegraphics[width=\linewidth]{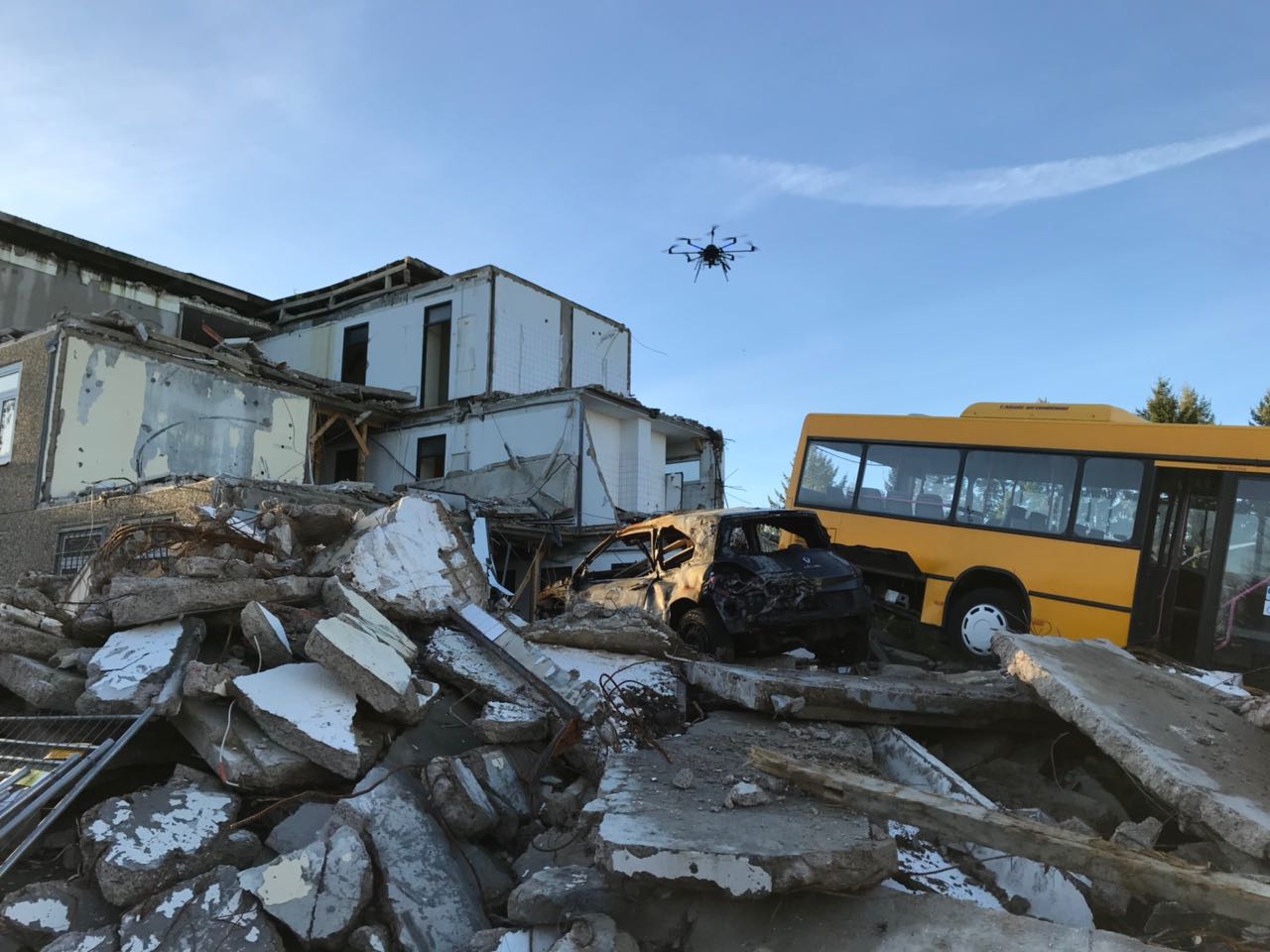}
    \caption{Real-World Outdoor Environment}
    \label{fig:outdoor_envmt}
  \end{subfigure}
  \caption{Evaluation scenarios for our autonomous landing system showing collapsed buildings, overturned vehicles and uprooted trees.}
  \label{fig:compare_envmt}
  \vspace{-0.5cm}
\end{figure}

We evaluate our system extensively in both simulated and real-world scenarios with collapsed buildings. We created a small city-scale simulation environment using the Unreal Engine consisting of collapsed buildings, damaged roads, debris and rubble piles, overturned cars, uprooted trees and several other features resembling a catastrophe-struck environment. While, for real-world evaluations, we exhaustively performed experiments at the TCRH Training Center for Rescue in Germany, spanning an area of 60,000 square meters and consisting of scenarios with earthquake and fire damage. \figref{fig:compare_envmt} shows example scenes from both these environments.

\subsection{Hyperrealistic Simulation Experiments}

\begin{figure*}
\footnotesize
\centering
\setlength{\tabcolsep}{0.1em} 
\begin{tabular}{p{0.5cm} p{0.9\textwidth}} 
\rot{(a) Hyperrealistic Simulation Environment} & \includegraphics[width=\linewidth]{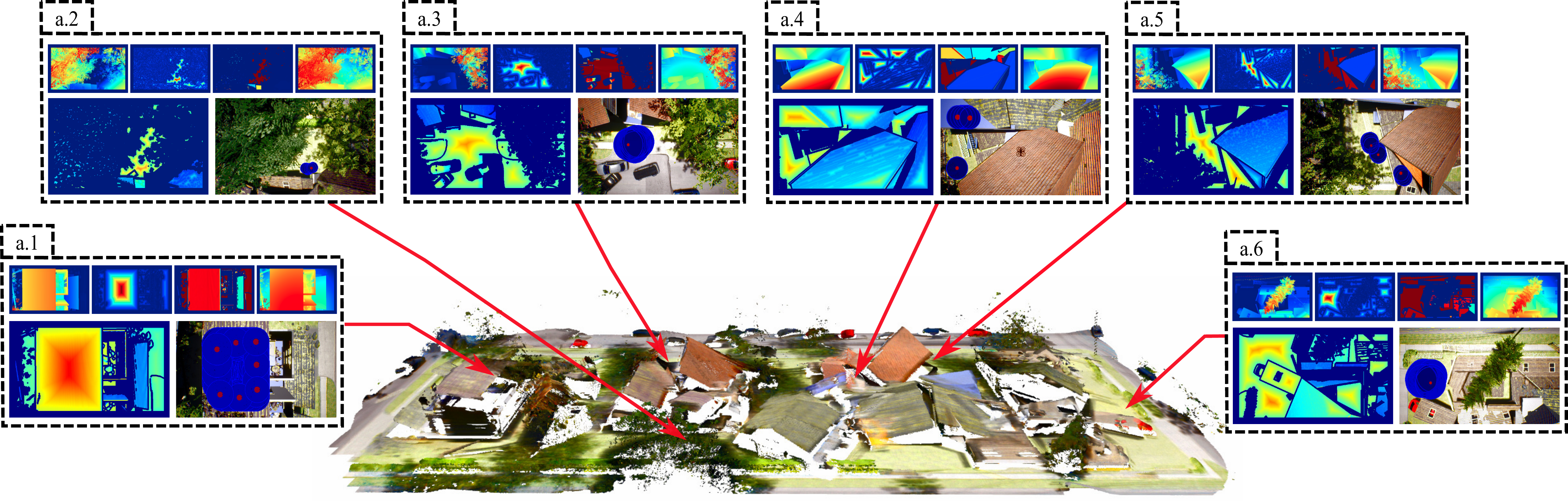} \\
\rot{(b) Real-World Outdoor Environment} & \includegraphics[width=\linewidth]{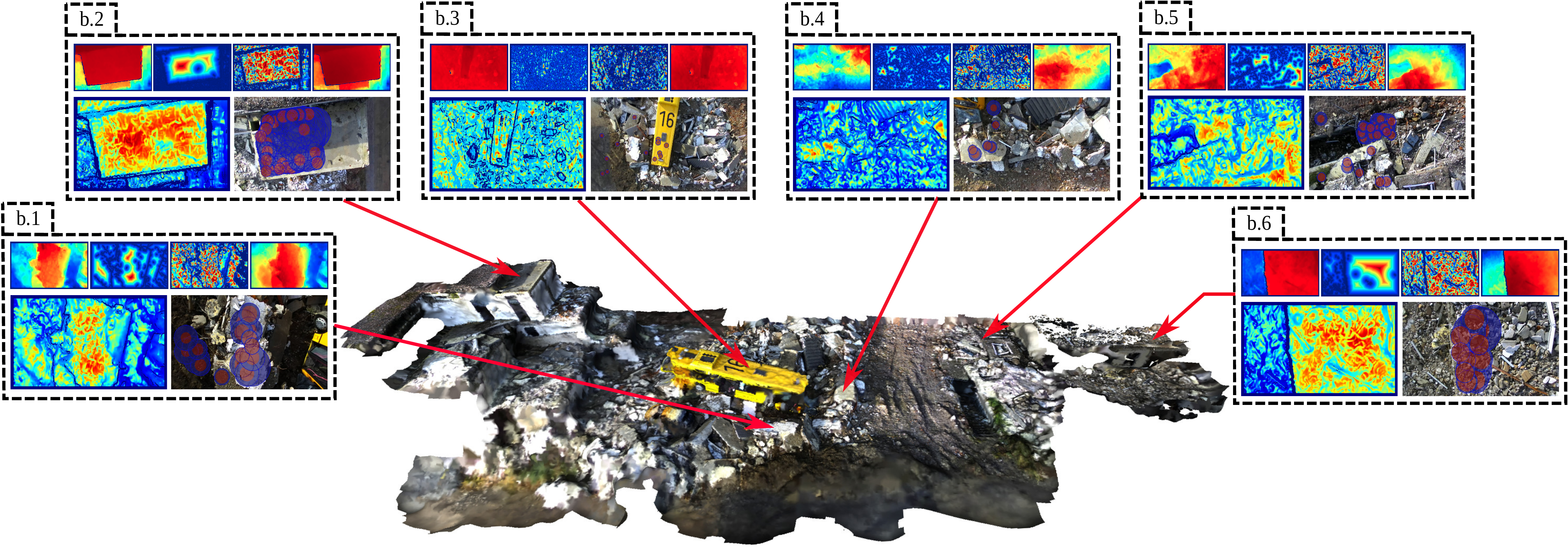} \\
\end{tabular} 
\caption{Illustrations of costmap evaluation and dense landing sites detection steps in both simulated and real-world scenarios. \textit{Inset:} The top row shows the depth accuracy, flatness, steepness and energy consumed costmaps respectively, while lower row shows the final decision map and the detected landing sites projected on to the camera image.}
\label{fig:results}
\vspace{-0.5cm}
\end{figure*}

We use the open-source \textit{AirSim} plugin~\cite{airsim2017fsr} with our Unreal Engine environment and ROS as the middleware for our simulation experiments. The simulator considers forces such as drag, friction and gravity in its physics engine, and provides data from various inertial sensors that are required for state estimation. We simulate a downward facing stereo camera mounted to the UAV which provides RGB-D data at $20\hertz$ with a resolution of $640\times480$ pixels, similar to our real-world setup. We clip the groundtruth depth map so that the depth sensor has the range $d_{min} = 0.05\meter$ and $d_{max} = 20.0\meter$. We use the same pipeline for state estimation, trajectory planning and landing site detection as in our real-world UAV system. We build an OctoMap at a resolution of $0.5\meter$ and the textured mesh at a resolution of $0.1\meter$. \figref{fig:results}(a) shows an example mesh visualization of a city block created using TSDFs while the UAV followed a lawn-mover surveillance path with a speed of $0.5\meter\per\second$.

\begin{figure}
\centering
\includegraphics[width=0.85\linewidth]{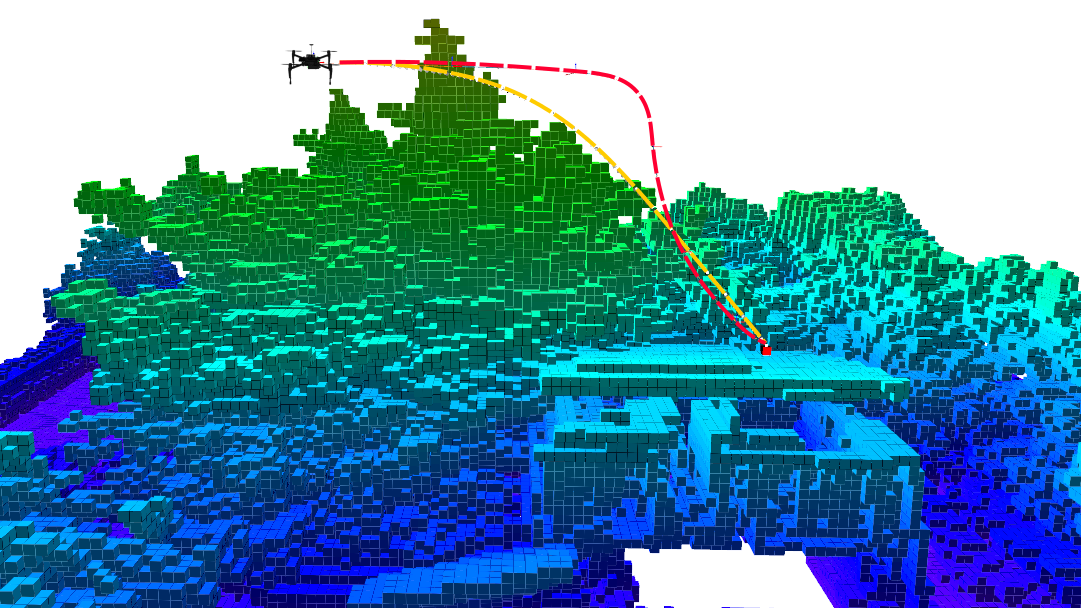}
\caption{Comparison of the path generated using a strictly sampling-based RRT* approach with a polynomial steer function (yellow line) and joint polynomial optimization method (red line). The landing site is selected from view $a.1$ in \figref{fig:results}. The polynomial-based steering function requires more time to compute a smooth path and has a higher cost than the trajectory generator that we use.}
\label{fig:compare_traj}
\vspace{-0.5cm}
\end{figure}

For the simulation experiments, we set the weights $c_1 = 0.05$, $c_2 = 0.4$, $c_3 = 0.4$ and $c_4 = 0.15$ for the depth accuracy, flatness, steepness and energy consumption costmaps respectively. We consider all the points with scores above $0.72$ in the final decision map as candidate landing sites. We set $c_1 = 0.05$ as the depth map is the groundtruth generated from the simulator. We choose equal weights for flatness and steepness costmaps since they play an equally important role in the landing sites detection. These two costmaps are critical for estimating safe landing sites in collapsed structures as they contain many parts of broken buildings piled on top of each other which appear as several cluttered planes in different orientations. We set the euclidean distance and depth threshold for hierarchical clustering of landing sites to $0.50\meter$ and $0.01\meter$. As it can be seen in the $a.1$ and $a.3$ scenarios, the flatness score ensures that the landing sites are not detected close to the edges of the roof of broken buildings or other obstructions. On the other hand, in the $a.4$ and $a.5$ scenarios, the steepness score that accounts for the orientation of the UAV, ensures that steep flat surfaces such as sections of broken walls are avoided from being detected as landing sites. Moreover, in the $a.2$, $a.3$, and $a.5$ scenarios, we can see that the trees are automatically given a lower score in both the flatness and steepness costmaps, thus discarding them from the landing site search area, which is crucial in catastrophe-struck environments as they often have uprooted trees. Fig.~\ref{fig:compare_traj} compares the trajectories generated using a strictly sampling-based RRT* approach and the joint polynomial optimization method used in our system.

\subsection{Real-World Outdoor Experiments}

We use the DJI M100 quadrotor equipped with a NVIDIA TX2 embedded computer and a downward facing ZED stereo camera for our real-world experiments. The stereo camera provides RGB-D images with a resolution of $640\times480$ pixels at $20\hertz$. As shown in \figref{fig:overview_system}, we use an EKF to fuse the raw inertial sensor data with GPS measurements and the pose output obtained from ORB-SLAM2. Similar to the simulation experiments, we use ROS as the middleware on the TX2 which runs all the processes in real-time for state estimation, trajectory planning, landing site detection and bioradar processing. We use an OctoMap resolution of $0.5\meter$ and a textured mesh resolution of $0.1\meter$. \figref{fig:results}(b) shows an example mesh visualization created by the UAV while exploration.\looseness=-1

Since in the real world, the depthmap generated from stereo images are often noisy, we set $c_1 = 0.15$ in order to account for this factor in the depth accuracy costmap. We set the weights for the flatness, steepness, and energy consumption costmaps to $c_2 = 0.35$, $c_3 = 0.4$ and $c_4 = 0.1$ respectively, while we set the overall thresholding value for the final decision map to $0.7$. For the hierarchical clustering parameters, we choose $0.5\meter$ for the euclidean distance (frame size of UAV being $0.26\meter$) and the depth threshold as $0.05\meter$. The effect of noisy depth data can also be seen in the other costmaps due to which the final decision map does not appear as clear as in the simulation environment. Nevertheless, our landing site search algorithm demonstrates detection of safe landing sites reliably even in situations where an expert safety operator is unable to make decisions. In scenarios $b.1$, $b.4$ and $b.5$, landing sites are clearly detected on flat surfaces engulfed by debris from collapsed buildings. Similar to the results observed in the simulation environment, several landing sites are reliably detected on roofs of collapsed structures in the $b.1$ and $b.2$ scenarios. Moreover, landing sites are also detected on the roof of the bus in the $b.3$ scenario. While, in the $b.6$ scenario, landing sites are detected closer to the edge of the roof containing small rubbles that have tolerable roughness for landing. These sites are often more safe to land on in comparison to other sites on the roof where larger rocks can be seen.\looseness=-1

\subsection{Computation Costs}

\begin{table}
\centering
\setlength\tabcolsep{0.32cm}
\caption{Runtime and memory consumption for processing each frame in the real-world scenario. Values are reported as $\mu \pm \sigma$. Evaluated on a system containing an Intel Core i7-8750H @ $2.20\giga\hertz$ CPU.}
\label{tab:timing}
\begin{tabular*}{8.5cm}{
  l
  S[table-format=-1.1(1)] S[table-format=-1.1(1)]
  S[table-format=-1.1(1)]
}
\noalign{\smallskip}\hline\noalign{\smallskip}
\textbf{Algorithm} & \textbf{Time ($\milli\second$)} & \textbf{Memory ($\mega\byte$)} \\
\noalign{\smallskip}\hline\hline\noalign{\smallskip}
   Costmap Evaluation &  &  \\
   \tab Depth Accuracy Costmap & 1.7\pm 2.4 & 20.7\pm 6.7 \\ 
   \tab Flatness Costmap &  91.7\pm 38.8 & 22.2 \pm 6.8\\ 
   \tab Steepness Costmap & 27.4\pm 4.2 & 69.3\pm 0.1\\ 
   \tab Energy Costmap & 1.6\pm 2.4 & 21.6\pm 5.5 \\ 
   \tab Final Costmap & 1.1\pm 2.1 & 21.3\pm 5.8\\
   Dense Landing Sites Detection & 15.5\pm 24.3 & 13.1\pm 0.2 \\ 
   Clustering & 28.5\pm 22.3 & 25.5\pm 0.2 \\
\noalign{\smallskip}\hline\noalign{\smallskip}
   \textbf{Total} & $\bf{167.5\pm 61.4}$ & $\bf{193.8\pm 6.6}$ \\ 
\noalign{\smallskip}\hline\noalign{\smallskip}
\end{tabular*}
\vspace{-0.5cm}
\end{table}

Runtime efficiency is one of the critical requirements of our system as all the on-board processes for state estimation, planning, landing site detection and bioradar analysis have to run online on the embedded computer. The total computation time of the entire landing site detection algorithm, with the OctoMap and textured mesh reconstruction along with the state estimation running in the backend, amounts to $167.5\milli\second$ and has a memory consumption of $193.8\mega\byte$. A detailed breakdown of this computation cost for each of the components of our algorithm is shown in \tabref{tab:timing}.

\begin{figure}
\footnotesize
\centering
  \begin{subfigure}[b]{\linewidth}
  \begin{center}
    \includegraphics[width=0.75\linewidth]{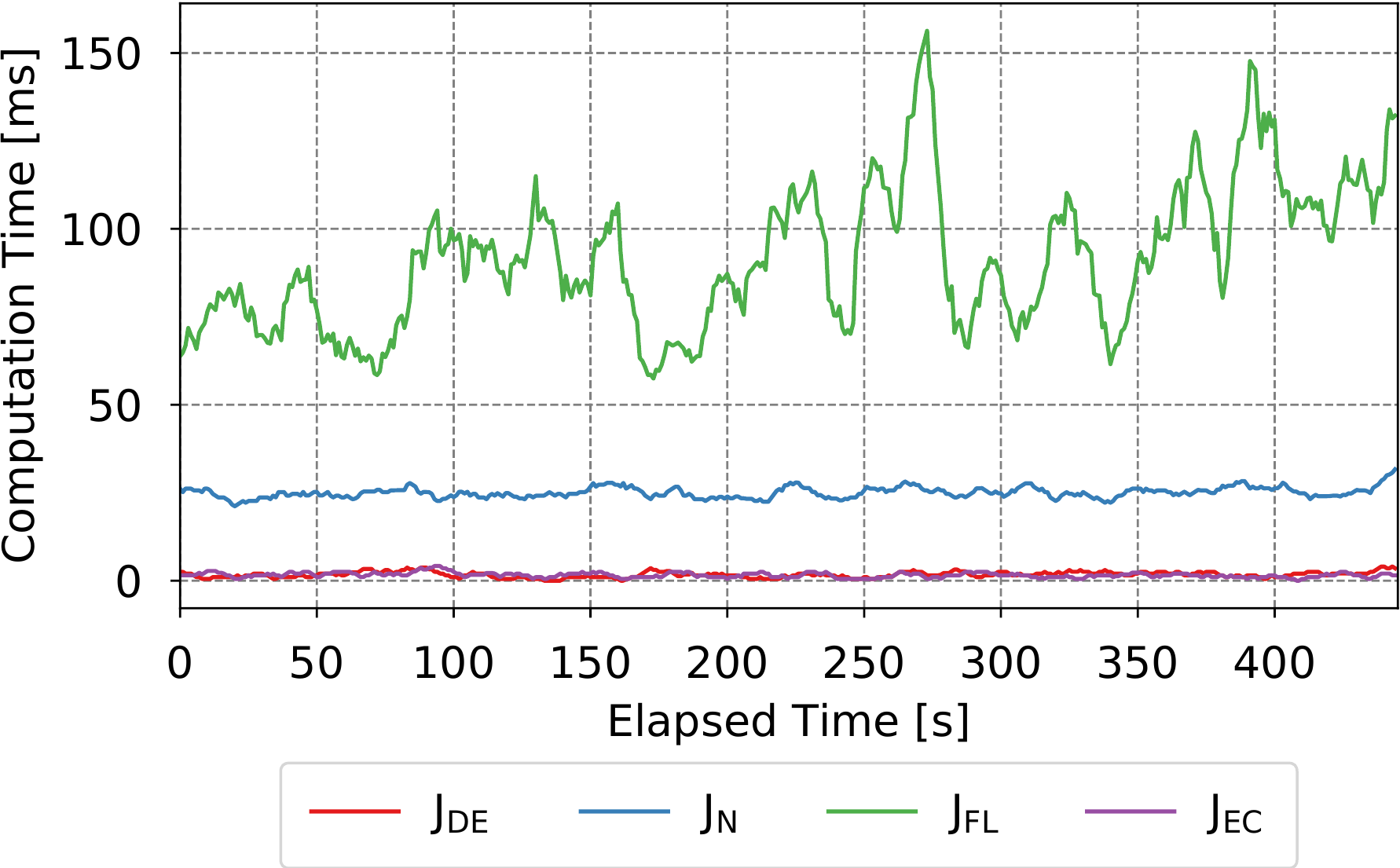}
    \caption{Computation time consumed to evaluate individual costmaps for each frame.}
    \label{fig:costmapT}
  \end{center}
  \end{subfigure}
  \vspace{0.1cm}
  
  \begin{subfigure}[b]{\linewidth}
  \begin{center}
  	\includegraphics[width=0.75\linewidth]{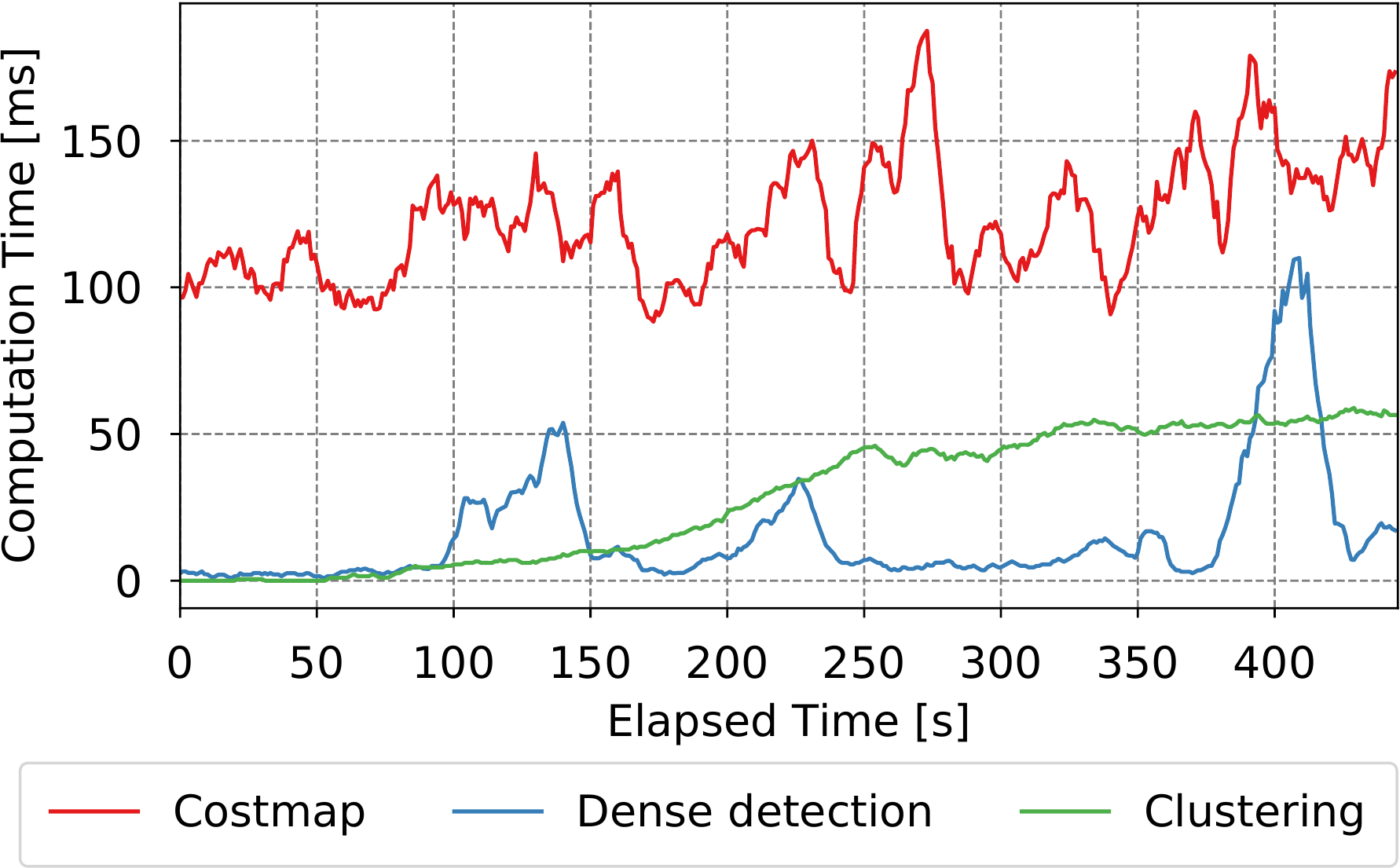}
  	\caption{Computation time consumed by each of the stages of the landing site detection.}
  	\label{fig:totalT}
  \end{center}
  \end{subfigure}
  \caption{Computational runtime analysis of our landing sites detection system. Number of samples considered: 455. Evaluated on Intel Core i7-8750H CPU @ 2.20GHz.}
  \label{fig:timing}
  \vspace{-0.5cm}
\end{figure}

The computation time for costmaps is evaluated for each input depth map. The cost of computing the depth accuracy $J_{DE}$, steepness $J_N$ and energy consumption $J_{EC}$ scores linearly depends on the size of the input image. During operation, the resolution of sensor images is unchanged leading to a constant computation time to evaluate these costmaps. On the other hand, the time consumed for evaluating the flatness score $J_{FL}$ depends on the distance transformation operation which varies with the image content of the binary map obtained from the canny edge operation. Due to this factor, a large variation in the computation time is observed while evaluating the flatness costmap. As shown in Fig.~\ref{fig:costmapT}, calculation of the flatness scores is the most time consuming step. It has a time complexity of $\mathcal{O}(dk)$ where $d=2$ for an image and $k$ is total number of pixels in the image.\looseness=-1

The dense landing sites detection step involves identifying candidate landing sites on the basis of their scores in the final decision map and aggregating these sites into a global list using k-d trees while removing duplicates. Since the number of landing sites detected in each frame varies according to the scene, we observe a large variance in the computation time for this step. However, for clustering, the time and memory complexity grow as $\mathcal{O}(n^2)$ and $\mathcal{O}(n)$ respectively with $n$ as the total number of landing sites detected.

\section{Conclusion}
\label{sec:conclusion}

In this paper, we presented a vision-based autonomous landing system for UAVs equipped with bio-radars tasked with search and rescue operations in catastrophe-struck environments. Our landing site detection algorithm considers several hazardous terrain factors including the flatness, steepness, depth accuracy and energy consumption information to compute a weighted costmap based on which we detect dense candidate landing sites. Subsequently, we employ nearest neighbor filtering and clustering to group dense sites into a safe landing region. We generate a low-resolution 3D volumetric map for trajectory planning and a high-resolution mesh reconstruction for structural analysis and visualization of the landing sites by the rescue team. We employ a polynomial trajectory planner to compute a minimum-jerk path to the landing site considering nearby obstacles and the UAV dynamics. Our proposed system is computationally efficient as it runs online on an on-board embedded computer with other processes for state-estimation and bioradar processing being run in the background. We demonstrated the utility of our system using extensive experiments in a hyperrealistic small city-scale simulation environment and in real-world environments with catastrophe-struck scenarios such as earthquakes and gas explosions.

\footnotesize
\bibliography{references.bib}
\bibliographystyle{unsrt}






\end{document}